\documentclass[letterpaper]{article} 
\usepackage{aaai24}  
\usepackage{times}  
\usepackage{helvet}  
\usepackage{courier}  
\usepackage[hyphens]{url}  
\usepackage{graphicx} 
\usepackage{natbib}  
\usepackage{caption} 
\usepackage{algorithm}
\usepackage[noend]{algpseudocode}
\usepackage{newfloat}
\usepackage{listings}
\DeclareCaptionStyle{ruled}{labelfont=normalfont,labelsep=colon,strut=off} 
\floatstyle{ruled}
\newfloat{listing}{tb}{lst}{}
\floatname{listing}{Listing}
\usepackage{bbding}
\usepackage{pifont}
\usepackage{subfigure}
\usepackage{epsfig}
\usepackage{amsmath}
\usepackage{amssymb}
\usepackage{enumitem}
\usepackage{booktabs}

\DeclareUrlCommand\url{\color{magenta}\def\UrlLeft{https://}\urlstyle{tt}}

\usepackage{xcolor}

\title{SMILEtrack: SiMIlarity LEarning for Occlusion-Aware Multiple Object Tracking}
\author{
    Yu-Hsiang Wang\textsuperscript{\rm 1}, Jun-Wei Hsieh\textsuperscript{\rm 1}*, Ping-Yang Chen\textsuperscript{\rm 2}, Ming-Ching Chang\textsuperscript{\rm 3}, Hung-Hin So\textsuperscript{\rm 4}, Xin Li\textsuperscript{\rm 3}\\
}
\affiliations{
    \textsuperscript{\rm 1}College of Artificial Intelligence and Green Energy, National Yang Ming Chiao Tung University, Taiwan.\\
    \textsuperscript{\rm 2}Department of Computer Science, National Yang Ming Chiao Tung University Taiwan.\\
    \textsuperscript{\rm 3}Department of Computer Science, University at Albany - SUNY.\\
    j122333221@gmail.com, [jwhsieh, pingyang.cs08]@nycu.edu.tw, mchang2@albany.edu, Xin.Li@mail.wvu.edu
}

\usepackage{bibentry}

\begin{document}

\maketitle

\begin{abstract}
Despite recent progress in Multiple Object Tracking (MOT), several obstacles such as occlusions, similar objects, and complex scenes remain an open challenge. Meanwhile, a systematic study of the cost-performance tradeoff for the popular tracking-by-detection paradigm is still lacking. This paper introduces SMILEtrack, an innovative object tracker that effectively addresses these challenges by integrating an efficient object detector with a Siamese network-based Similarity Learning Module (SLM). The technical contributions of SMILETrack are twofold.
First, we propose an SLM that calculates the appearance similarity between two objects, overcoming the limitations of feature descriptors in Separate Detection and Embedding (SDE) models. The SLM incorporates a Patch Self-Attention (PSA) block inspired by the vision Transformer, which generates reliable features for accurate similarity matching. Second, we develop a Similarity Matching Cascade (SMC) module with a novel GATE function for robust object matching across consecutive video frames, further enhancing MOT performance. Together, these innovations help SMILETrack achieve an improved trade-off between the cost ({\em e.g.}, running speed) and performance ({\em e.g.}, tracking accuracy) over several existing state-of-the-art benchmarks, including the popular BYTETrack method. SMILETrack outperforms BYTETrack by {\bf 0.4-0.8 MOTA} and {\bf 2.1-2.2 HOTA} points on MOT17 and MOT20 datasets. Code is available at {\footnotesize \url{github.com/pingyang1117/SMILEtrack_official}}.
\end{abstract}


\section{Introduction}

The task of Multiple Object Tracking (MOT) is to estimate the trajectories of each target and associate them between frames in video sequences. MOT has found widespread applications in various fields, including computer interaction~\cite{6618967, luo2017learning}, smart video analysis, and autonomous driving. Modern MOT systems \cite{2016, wang2019towards} typically follow the Tracking-By-Detection (TbD) paradigm, which involves two separate steps of detection and tracking. The detection step locates the object of interest in a single video frame, while the tracking step links each detected object to the existing tracks or creates new tracks if none are found. Despite enormous efforts in MOT investigation, the task remains challenging due to vague objects, occlusion, and complex scenes in real-world applications.

\begin{figure}[t]
\centerline{
\includegraphics[width=\linewidth]{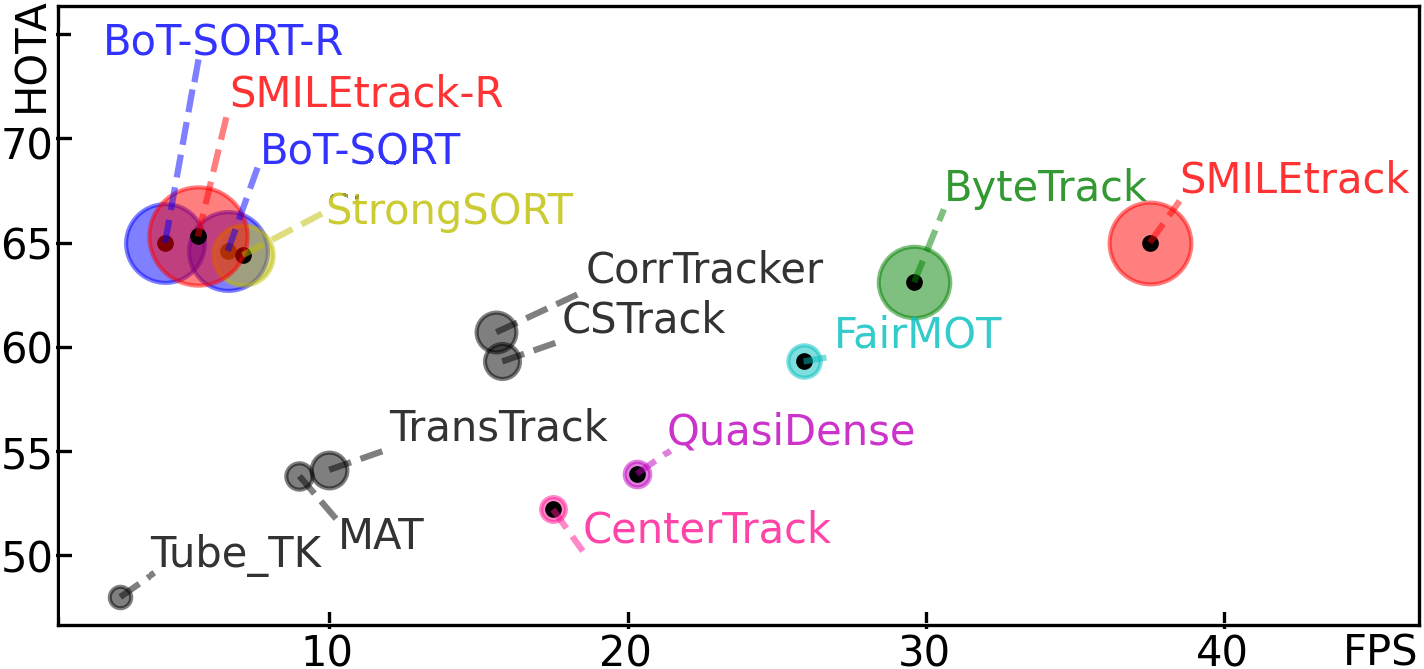}
\vspace{-2mm}
}
\caption{ 
Comparative analysis of HOTA-MOTA-FPS for different trackers on the MOT17 test set. X-axis: FPS (running speed). Y-axis: HOTA. Circle radius: MOTA score. SMILEtrack registers 80.7 MOTA and 65.0 HOTA at 37.5 FPS, exceeding all other trackers (see Table~\ref{tab:SoTA:MOT17} for details).
}
\label{fig:teaser}
\vspace{-4mm}
\end{figure}

\begin{figure*}[t]
\centerline{
\includegraphics[width=\linewidth]{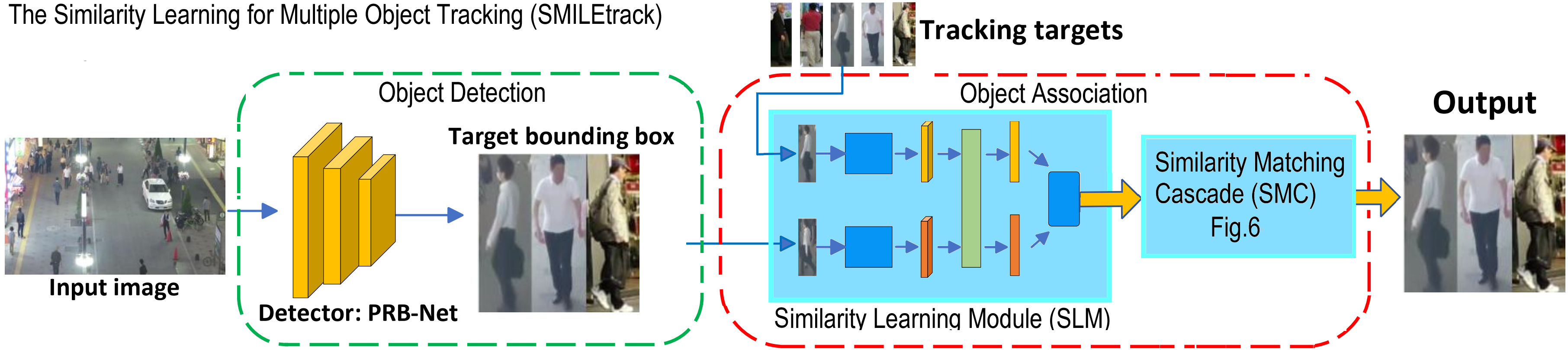}
\vspace{-2mm}
}
\caption{ 
The architecture of the proposed SMILEtracker. SMILEtracker is a Siamese network-like architecture that learns the appearance features of two objects and calculates their similarity score.  SMILEtracker consists of two modules: (i) object detection and (ii) object association.  
}
\label{SMILT_Tracker}
\vspace{-4mm}
\end{figure*}

In the Tracking-By-Detection (TbD) paradigm, two primary strategies prevail, namely Joint Detection and Embedding (JDE) and Separate Detection and Embedding (SDE). JDE methods~\cite{wang2019towards, zhang2021fairmot} combine the detector and the embedding model into a single-shot deep network that outputs the detection results and the corresponding appearance embedding features in one inference.  
Alternatively, SDE methods~\cite{2016,du2023strongsort, aharon2021botsort} require a detector and a re-identification model. The detector locates all objects in a single frame via bounding boxes~\cite{ren2016faster,ssd2016,redmon2016look, 8100173, redmon2018yolov3, yolov4,prb}. The re-identification model then extracts the embedding features of each object from its bounding box, and these features are used to associate each bounding box with one of the existing trajectories. Despite their flexibility, the efficiency of SDE methods trails behind that of JDE due to the necessity of two separate models.
The Tracking-by-Attention (TbA) paradigm~\cite{zhang2021trackformer,peng2021transtrack,yang2021motsnet,li2021set} applies attention to data associations and jointly performs tracking and detection via Transformer~\cite{attent}. 

The motivation behind this work is two-fold. One of the long-standing problems in MOT is occlusion handling, and the other is a principled solution to speed-accuracy trade-off.
Although the TbA method has impressive results on feature attention, its exceptional feature attention results in a high time complexity that reduces inference speed. In addition, occlusions can cause tracked objects to pay less attention, resulting in the failure of MOT. Meanwhile, TbD methods such as ByteTrack~\cite{zhang2021bytetrack} enjoy computational efficiency, but their accuracy is not optimized. It is highly desirable to develop a class of MOT methods that can strike an improved trade-off between cost ({\em e.g.}, running speed measured by FPS) and performance ({\em e.g.} tracking accuracy measured by MOTA~\cite{Bernardin2018Metrics}).


This paper proposes a novel object tracker, {\bf Similarity Learning for Multiple Object Tracking (SMILEtrack)}, which combines an object detector and a Similarity Learning Module (SLM) to address various challenges in MOT, especially occlusion.  Fig.~\ref{SMILT_Tracker} shows the architecture of our SMILEtrack, which provides two major contributions to achieving the State-of-the-Art (SoTA) MOT system: (1) an efficient and lightweight self-attention mechanism that learns the similarity between two candidate bounding boxes. Although the SDE model can achieve high accuracy in object tracking, most feature descriptors used in the model cannot differentiate between objects with similar appearances. To solve this problem, we propose a Siamese network-based Similarity Learning Module (SLM) that can calculate the similarity in appearance between two objects. Inspired by the vision Transformer~\cite{vit}, we introduce a Patch Self-Attention (PSA) block in SLM to produce reliable features for similarity matching. (2) a robust tracker with a novel GATE function that can associate each candidate bounding box from video frames, leading to improved MOT performance.  To better handle occlusions, we create a Similarity Matching Cascade (SMC) module that takes SLM results and matches multiple objects robustly across frames. The proposed network achieves SoTA performance on the MOT17 and MOT20 datasets. Contributions of our work are summarized as follows.

\begin{itemize}[leftmargin=8pt]\itemsep -.1em
\item We propose SMILETrack, a separate detection and tracking model, to track multiple objects in frames. SMILETrack can outperform BYTETrack \cite{zhang2021bytetrack} by 0.4-0.8 MOTA points and over 2.0 HOTA points on the MOT17 and MOT20 datasets; see to Fig.~\ref{fig:teaser}.

\item We introduce a Siamese network-based Similarity Learning Module (SLM) to learn the similarity in appearance between objects for tracking.

\item A Patch Self-Attention (PSA) block is proposed that uses a self-attention mechanism to produce reliable features for similarity matching.

\item We design a Similarity Matching Cascade (SMC) module to match objects more reliably across frames, which improves performance largely in the presence of occlusions. 
\end{itemize}

\section{Related Work}

\subsection{Tracking-by-Detection}
\label{related:TbD}

The Tracking-by-Detection (TbD) method has become one of the most popular approaches in the MOT framework. The main tasks of the TbD method can be roughly divided into two parts: object detection and object association. 

\textbf{Object Detection}: Mainstream visual object detection models fall into two categories, namely, the two-stage (proposal-driven) and one-stage (direct) detectors. The two-stage methods~\cite{ren2016faster} offer high accuracy but at the cost of speed. On the contrary, one-stage methods are faster but less accurate. 
YOLO object detection models~\cite{redmon2016look, 8100173, redmon2018yolov3, yolov4}  have been widely used in multi-object tracking (MOT) applications due to their speed and accuracy. However, these anchor-based detectors introduce many hyperparameters and consume significant time and memory during training. To mitigate these issues, anchor-free detectors such as CenterNet~\cite{zhou2019objects}, and YOLOX~\cite{yolox} have emerged. Despite their improvements~\cite{zhang2021bytetrack,aharon2021botsort}, these tracking devices still struggle to accurately detect objects of varying sizes. PRB-Net~\cite{prb} is an effective object detector for MOT tasks, addressing the limitations of anchor-based and anchor-free detectors. 

\textbf{Object Association}: 
SORT~\cite{2016} is a simple effective tracking algorithm that uses Kalman filtering and Hungarian matching for object association. It struggles with challenges such as occlusions and fast-moving objects. DeepSORT~\cite{8296962} alleviates occlusion issues by incorporating CNN-based appearance features; however, this compromises execution speed.
To address this efficiency issue, FairMOT~\cite{zhang2021fairmot} employs an anchor-free method based on CenterNet~\cite{zhou2019objects}, which significantly improves the MOT performance on the MOT17 dataset.
To improve tracking efficiency, numerous MOT methods~\cite{StadlerBeyerer2021_1000141653, Stadler_2022_WACV} ignore the appearance features of objects, instead leveraging high-performance detectors and motion cues. Despite achieving impressive results and fast inference on MOTChallenge~\cite{milan2016mot16} benchmarks, we posit that their performance is largely dependent on the simplicity of the movement patterns of the dataset. Omitting appearance features may compromise tracking accuracy and robustness in densely populated scenes.

\subsection{Tracking-by-Attention}

Trackformer~\cite{meinhardt2021trackformer} extends its success in object detection to MOT by casting the task into a frame-to-frame set prediction problem. Data association between frames is calculated through attention, and a set of track predictions across frames is evolved using the encoder-decoder architecture of Transformer. Similarly, TransTrack~\cite{transtrack:arxiv:2020} uses an attention-based query-key mechanism to perform object detection and association in a single shot based on Deformable DETR~\cite{DefDetr:ICLR:2021}. TransCenter~\cite{TransCenter:TPAMI:2021} is another Transformer-based architecture that uses image-related dense detection queries and sparse tracking queries for MOT. However, all Transformer-based schemes are computationally intensive, and thus not suitable for real-time applications.

\section{Methodology}


We introduce {\bf Similarity Learning for Multiple Object Tracking (SMILEtrack)}, a novel MOT architecture integrating a detector~\cite{prb} and a Similarity Learning Module (SLM). SMILEtrack comprises two modules, as shown in Fig.~\ref{SMILT_Tracker}: {\em object detection} and {\em object association}. The former model was designed primarily to excel in localizing large and small pedestrians, achieving both accuracy and efficiency, making it a superior choice over YOLOX~\cite{yolox}.
The technical contributions of this work are mainly in the latter module, which consists of: (1) {\em similarity calculation}, where a novel similarity learning module (SLM) learns the appropriate features and computes an appearance affinity matrix using a Siamese network; and (2) {\em object association}, where a Similarity Matching Cascade (SMC) module solves the MOT linear assignment problem using the Hungarian algorithm. Details are explained in the following sections.




\subsection{Similarity Learning Module (SLM)}

Object appearance information is essential for achieving robust tracking quality. Although SORT is a simple association framework that can achieve high-speed inference time, its similarity score does not consider object appearance information and cannot handle long-term occlusion or objects with fast motion. DeepSORT~\cite{8296962} addresses this problem by using a pre-trained CNN to compute bounding-box appearance descriptors. However, this descriptor only considers the similarity between the same objects, without considering the dissimilarity between different objects in different frames. Here, we propose the Similarity Learning Module (SLM) that leverages a Siamese network architecture to learn more discriminative appearance features and accurately track objects across frames. 

\begin{figure}[t]
\centerline{
\includegraphics[width=\linewidth]{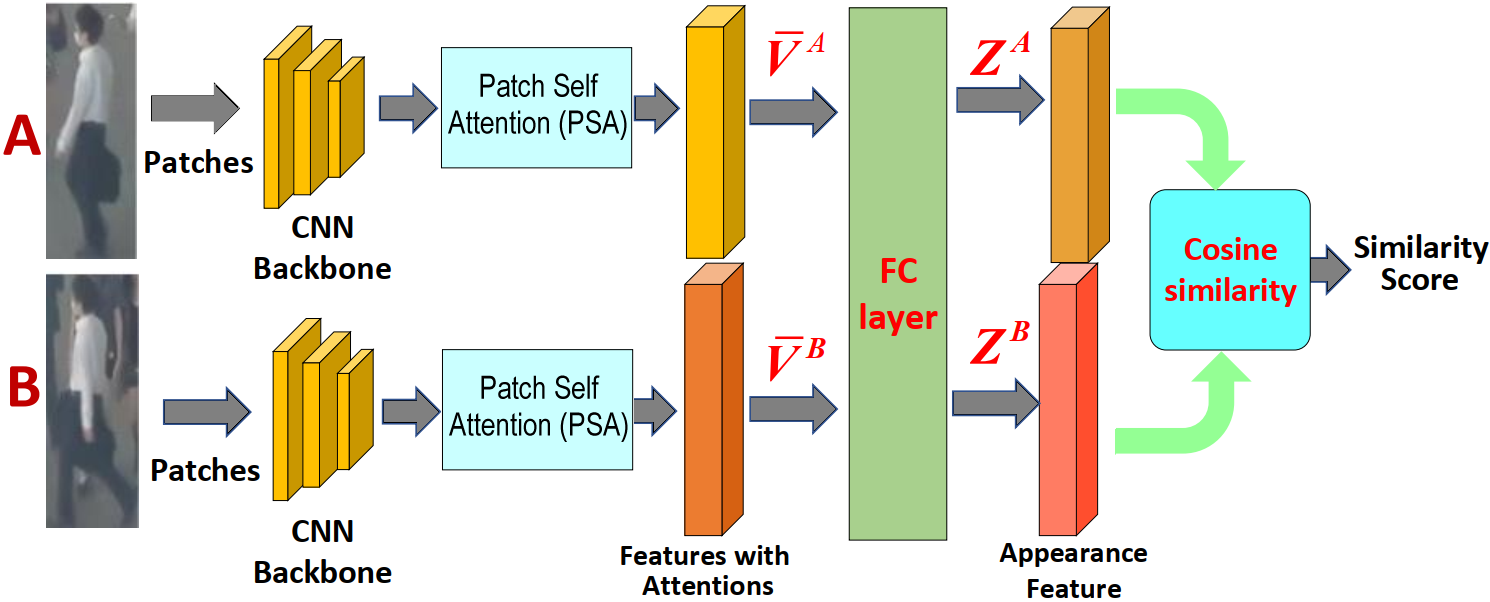}
\vspace{-2mm}
}
\caption{ 
Appearance similarity between low-score detection at the current frame and tracks at the previous frame.
}
\label{fig:SLM}
\vspace{-4mm}
\end{figure}

\begin{figure}[t]
\centerline{
\includegraphics[width=\linewidth]{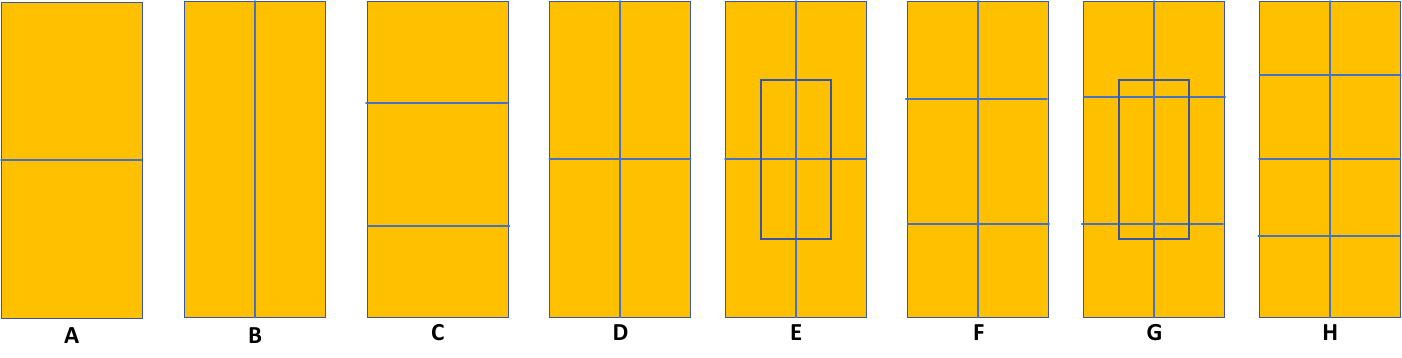}
\vspace{-2mm}
}
\caption{Different types of patch layout: configuration (E) achieves the best performance because it can actively attend to PSA-occluded parts when occlusion occurs. }
\label{PatchLayout}
\vspace{-4mm}
\end{figure}

Fig.~\ref{fig:SLM} shows the SLM architecture. It takes the target and query objects as input in the Siamese network. Both are divided into several patches and then pass through the {\bf Patch Self-Attention (PSA)} block. Note that the height-width ratio of all patches is not fixed (see Fig. \ref{PatchLayout}). Since objects of interest in the MOT17 and MOT20 datasets are assumed to be pedestrians, we have found that configuration (E) achieves the best performance. This can be explained away by observing that layout (E) exploits both prior knowledge about walking pedestrians (i.e., the height-width ratio is approximately 2:1) and translation invariance (i.e., the center box is a shifted version of four surrounding boxes).  

\subsubsection{Patch Self-Attention (PSA) Block}
\label{sec:ISA}

\begin{figure}[t]
\centerline{
\includegraphics[width=0.85\linewidth]{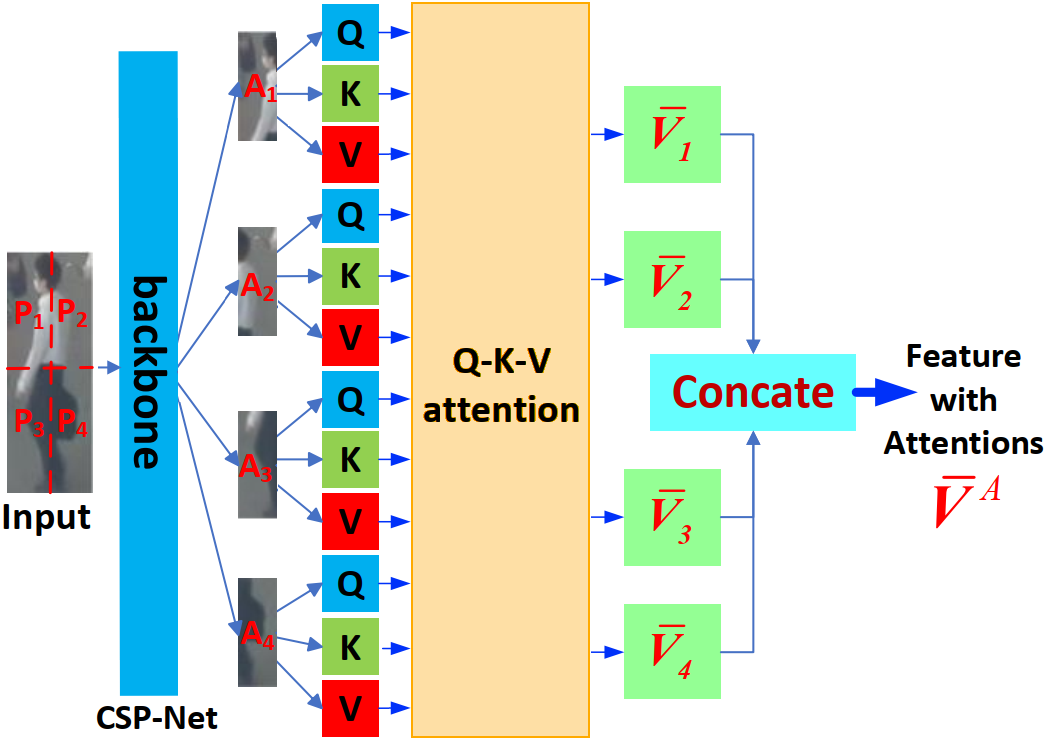}
\vspace{-2mm}
}
\caption{The Patch Self-Attention (PSA) architecture.}
\label{PSA_Block}
\vspace{-4mm}
\end{figure}

To produce a reliable appearance feature, a superior feature representation is essential. Inspired by the Vision Transformer (VIT)~\cite{vit}, each SLM input is divided into separate patches. Then, all the patches and their positions are embedded together and fed into a backbone to extract rich feature vectors.  Then, three fully connected networks are adopted to convert the deep visual features of all patches to three sets of compact features, $i.e.$, query, key, and value.  Based on the features from the query and key sets, various attentions among different combinations can be calculated and used to weight the features from the value set of each patch to form a feature vector to represent an object more accurately. The detailed architecture of the PSA block is shown in Fig.~\ref{PSA_Block}. 

\subsubsection{The Q-K-V Attention}
Since input objects are of different sizes, we resize them to a fixed size $W \times  H$ where  $W$ and $H$ are, respectively, set to 80 and 224 in this paper. Assume that an object $\boldsymbol{A}$ is divided into $N_P$ patches $\{P_i \}_{i=1,...,N_P}$. Each patch $P_i$ has a fixed size $W_P \times H_P$. Then, we use a row-major scanning order to convert each $P_i$ to a column vector.  Then, an object can be represented as a sequence of $N_p$ feature vectors: $ (P_1,...,P_i, ..., P_{N_p}), P_i \in R^{D_p}$,  where $D_p=W_p \times H_p$. 
Before feature extraction, the values of pixels in $P_i$ are normalized to $[0,1]$. Since there are geometrical relations between the patches in $\boldsymbol{A}$, their representation should be modified to preserve position-dependent properties. For the $i$th patch $P_i$, its position embedding vector $E_i$ is specified by the standard transformer \cite{vaswani2017attention}.
It follows that an object $\boldsymbol{A}$ is embedded as $\boldsymbol{A} = (A_1,...,A_i, ..., A_{N_P})$, where ${A}_i = P_i + E_i$ and $\boldsymbol{A} \in R^{D_p \times N_P}$. For each ${A}_i$, we adopt the CSP-Net framework~\cite{CSPNet:CVPRW2020} as the backbone to convert it into a feature matrix $\boldsymbol{F_i}$. 
$\boldsymbol{F_i}$ includes $d_f$ row vectors and $C$ column vectors; that is,  $\boldsymbol{F_i} \in R^{d_f \times C}$, where $C$ is the number of feature channels and $d_f$ is the size of the last layer of the feature pyramid created by CSP-Net~\cite{CSPNet:CVPRW2020}.  
 
Let $\boldsymbol{W}_Q$, $\boldsymbol{W}_K$, and $\boldsymbol{W}_V$ be three learned linear transforms that map $\boldsymbol{F_i}$ to the query $\boldsymbol{Q_i}$, the key $\boldsymbol{K_i}$, and the value $\boldsymbol{V_i}$, respectively. Assume that $\boldsymbol{W}_K$ and $\boldsymbol{W}_Q$ have the same number of column vectors, \emph{i.e.}, $d_k$. Also, there are $d_v$ column vectors in $\boldsymbol{W}_V$.  Then $\boldsymbol{W}_Q \in R^{C\times d_k}$, $\boldsymbol{W}_K \in R^{C\times d_k}$, and $\boldsymbol{W}_V \in R^{C\times d_v}$.  With $\boldsymbol{W}_Q$, $\boldsymbol{W}_K$, and $\boldsymbol{W}_V$, we can obtain $\boldsymbol{Q_i}$, $\boldsymbol{K_i}$, and $\boldsymbol{V_i}$ by the following equations: 
 \begin{equation}\label{eq:QKV} 
 \boldsymbol{Q_i} = \boldsymbol{F_i}\boldsymbol{W}_Q,  \boldsymbol{K_i}=\boldsymbol{F_i}\boldsymbol{W}_K, \boldsymbol{V_i}=\boldsymbol{F_i}\boldsymbol{W}_V,
\end{equation}
where  $\boldsymbol{Q_i} \in R^{d_f \times d_k}$, $\boldsymbol{K_i} \in R^{d_f \times d_k}$, and $\boldsymbol{V_i} \in R^{d_f \times d_v}$. Before matching, the norms of $\boldsymbol{Q_i}$, $\boldsymbol{K_i}$, and $\boldsymbol{V_i}$ will be normalized to be one; that is,  $||\boldsymbol{Q_i}||$=1, $||\boldsymbol{K_i}||$=1, and $||\boldsymbol{V_i}||$=1.

Let $\otimes$ denote the Hadamard product and $Sum(\boldsymbol{M})$ be an element-wise sum on a matrix $\boldsymbol{M}$. For $A_i$, its attention $\alpha_{i,j}$ to $A_j$ can be calculated according to the following equation: 
\begin{equation}\label{eq:Attentions} 
\alpha^{i,j} = \frac{Sum(\boldsymbol{Q}_i \otimes {\boldsymbol{K}_j )}}{\sum\limits_{j=1}^{N_p}{Sum(\boldsymbol{Q}_i \otimes {\boldsymbol{K}_j )}}}.
\end{equation}
With  $\alpha_{i,j}$,  $A_i$ is converted to a feature vector $\Bar{\boldsymbol{V}}_{i}$ as follows: $\Bar{\boldsymbol{V}}_{i} = {\sum\limits_{j=1}^{N_p}{\alpha_{i,j}{\boldsymbol{V}_j}}}$.
After concatenating all $\Bar{\boldsymbol{V}}_{i}$, a new feature vector ${\Bar{\boldsymbol{V}}}^A$ is created from $\boldsymbol{A}$ for object tracking: 
$ {\boldsymbol{\Bar{\boldsymbol{V}}}^A = (\Bar{\boldsymbol{V}}_1,...,\Bar{\boldsymbol{V}}_i, ..., \Bar{\boldsymbol{V}}_{N_P})}.$
In Fig.~\ref{PSA_Block}, after the PSA block, ${\Bar{\boldsymbol{V}}}^A$ is converted to a new feature vector $\boldsymbol{Z}^A$ by using a fully-connected network.  Then, given two objects $\boldsymbol{A}$ and $\boldsymbol{B}$, with SLM, their similarity score can be measured by calculating the cosine similarity between $\boldsymbol{Z}^A$ and $\boldsymbol{Z}^B$.

\begin{figure}[t]
\centerline{
\includegraphics[width=0.9\linewidth]{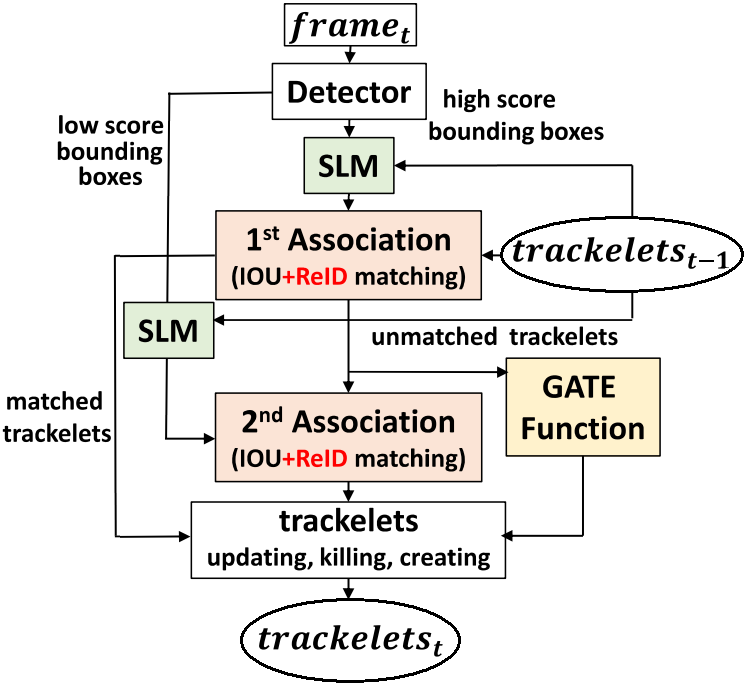}
\vspace{-2.5mm}
}
\caption{ 
The Similarity Matching Cascade (SMC) pipeline.
}
\label{fig:SMC:pipeline}
\vspace{-4mm}
\end{figure}

\subsection{Similarity Matching Cascade (SMC) for Tracking} 


{\em Object association} is the crucial step after similarity calculation for MOT. A well-designed association strategy can have a significant impact on tracking results such as HOTA \cite{HOTA2021}. In the literature, ByteTrack~\cite{zhang2021bytetrack} is a simple yet effective method of association with objects, where detected boxes are classified by their confidence scores from high to low, and the best match in history is found based on the IOU criterion. Although ByteTrack achieves SoTA performance in some MOT evaluations (i.e., simple motion patterns), relying solely on the IOU distance for data association can result in frequent ID switches when visually similar targets approach each other (e.g., one occludes the other). To address this issue, we designed the SMC association method as shown in Fig.~\ref{fig:SMC:pipeline} that integrates the advantages of ByteTrack to achieve an improved trade-off between speed and accuracy.

Let $\mathbb{O}$ denote the set of objects detected by the PRB-Net from the current frame. All objects $O_i$ in $\mathbb{O}$ are sorted according to their detection scores in descending order (the median detection score is $\mu$). Subsequently, all objects $O_i$ in $\mathbb{O}$ are divided into two sets: $\mathbb{O}^H$ and $\mathbb{O}^L$-based thresholding. Any object in $\mathbb{O}$ with a detection score higher than the threshold $\mu$ is placed in $\mathbb{O}^H$. If its detection score is lower than $\mu$ but higher than 0.1, it belongs to $\mathbb{O}^L$. We treat an object as background or noise if its detection score is below $0.1$. Two different association strategies are employed to match elements in $\mathbb{O}^H$ and $\mathbb{O}^L$, respectively.

\begin{table*}[t]
\caption{Comparison against the SoTA MOT methods on the MOT17~\cite{milan2016mot16} test set.
\vspace{-2mm}
}
\label{tab:SoTA:MOT17}
\small
\setlength\tabcolsep{4.5pt}
\centerline{
\begin{tabular}{c|ccccccccc}
\toprule 
Method &
  MOTA $\uparrow$ &
  IDF1 $\uparrow$ &
  HOTA $\uparrow$ &
  FN $\downarrow$ &
  FP $\downarrow$ &
  IDs $\downarrow$ &
  MT $\uparrow$ &
  ML $\downarrow$ &
  FPS $\uparrow$ \\ \toprule 
Tube\_TK~\cite{tubetk:CVPR:2020} & 63.0 & 58.6 &48.0 & 177,483 & 27,060 & 4,137 & 31.2\% & 19.9\%  &3.0\\
MOTR~\cite{MOTR:ICCV:2021}  & 65.1  & 66.4 & - & 149,307 & 45,486 & 2,049 & 33.0\% & 25.2\%  &-\\
CTracker~\cite{CTracker:ECCV:2020} &66.6 &57.4 & - &160,491 &22,284 &5,529 &- &- &- \\
CenterTrack~\cite{CenterTrack:ECCV:2020}  & 67.8 & 64.7 & 52.2 & 160,332 & 18,498 & 3,039 & 34.6\% & 24.6\%  &17.5\\
QuasiDense~\cite{QuasiDense:CVPR:2021} &68.7  &66.3 & 53.9 & 146,643 & 26,589 & 3,378 & 40.6\% & 29.1\%  &20.3\\
TraDes~\cite{TraDes:CVPR:2021} &69.1 &63.9 &- &150,060 &20,892 &3,555 &- &-  &-\\
MAT~\cite{MAT:Neurocomputing:2021} &69.5 &63.1 & 53.8 &138,741 &30,660 &2,844 &43.8\% &18.9\%  &9.0\\
SOTMOT~\cite{Zheng_2021_CVPR} & 71.0 & 71.9 & - & 118,983 & 39,537 & 5,184 & 42.7\% & 15.3\% &16.0\\
GSDT~\cite{GSDT:ICRA:2021}& 73.2 & 66.5 & - & 120,666 & 26,397 & 3,891 & - & - &-\\
FairMOT~\cite{zhang2021fairmot} &73.7 &72.3 & 59.3 &117,477 &27,507 &3,303 &43.2\% &17.3\%  &25.9\\
RelationTrack~\cite{RelationTrack:ICIP:2021}&73.8 &74.7 & - &118,623 & 27,999 &1,374 &- &- &-\\
PermaTrackPr~\cite{PermaTrackPr:ICCV:2021} &73.8 &68.9 &- &115,104 & 28,998 &3,699 &- &- &-\\
CSTrack~\cite{CSTrack:TIP:2021} &74.9 &72.6 & 59.3 &114,303 &23,847 &3,567 &41.5\% &17.5\% &15.8\\
TransTrack~\cite{transtrack:arxiv:2020} &75.2 & 63.5 &54.1 & 86,442  & 50,157 & 3,603 & 55.3\% & \textbf{10.2}\% &10.0\\
SiamMOT~\cite{SiamMOT:AAAI:2021}&76.3 &72.3 &- &-  &- &- &- &- &-\\
TransCenter~\cite{TransCenter:TPAMI:2021} &76.4 &65.4 &- &89,712  &37,005 &6,402 &51.7\% &11.6\%  &-\\
CorrTracker~\cite{CorrTracker:CVPR:2021} &76.5 &73.6 &60.7 &99,510  &29,808 &3,369 &47.6\% &12.7\% &15.6\\
TransMOT~\cite{TransMot:ICCV:2021} &76.7 &75.1 &- &93,150  &36,231 &2,346 & - & - &-\\
ReMOT~\cite{ReMOT:IVC:2020} &77.0 &72.0 &- &93,612  &33,204 &2,853 & - & -  &-\\
OCSORT~\cite{cao2022observation} & 78.0 &77.5 &- & 107,055  & \textbf{15,129} & 1,950 & - & - &-\\
MAATrack~\cite{Stadler_2022_WACV} &79.4 &75.9 &62.0 &77,661 &37,320 &1,452 &- &-  &-\\
StrongSORT++~\cite{du2023strongsort}  & 79.6 & 79.5 & 64.4 & 86,205  & 27,876 & \textbf{1,194} &53.6\% & 13.9\%  &7.1\\
ByteTrack~\cite{zhang2021bytetrack} & 80.3 & 77.3 & 63.1 & 83,721  & 25,491 & 2,196 & 53.2\% & 14.5\%  &29.6\\
BoT-SORT~\cite{aharon2021botsort} & 80.6 & 79.5 & 64.6 & 85,398  & 22,524 & 1,257 & - & -  &6.6\\ 
\textbf{SMILEtrack w/o Re-ID (Ours)} & 80.7 & 80.1 & 65.0 & 81,792 & 23,187 & 1,251 & 54.7\% & 14.2\%  &\textbf{37.5}\\ \midrule
BoT-SORT-R~\cite{aharon2021botsort} & 80.5 & 80.2 & 65.0 & 86,037  & 22,521 & 1,212 & - & -  &4.5\\ 
\textbf{SMILEtrack (Ours)} & \textbf{81.1} & \textbf{80.5} & \textbf{65.3} & \textbf{79,428} & 22,963 & 1,246 & \textbf{56.3}\% & 14.7\%  &5.6\\ 
\bottomrule
\end{tabular}
}
\vspace{-4mm}
\end{table*}

Let $\mathcal{T}$ represent the track list stored in the previous frame. Before matching, each track in $\mathcal{T}$ predicts its new position in the current frame using a Kalman filter. Moreover, $\mathcal{T}_i(k)$ denotes the $k$th fragment or tracklet of the $i$th track in $\mathcal{T}$, where $\mathcal{T}_i(\text{last})$ refers to the last fragment of $\mathcal{T}_i$.  Furthermore, we use $S_{iou}^H(i,j)$ and $S_{app}^H(i,j)$ to denote the IOU similarity matrix and the appearance similarity matrix, respectively, between $\mathcal{T}_i(\text{last})$ and the $j$th object $O_j$ in $\mathbb{O}^H$. The value of $S_{app}^H(i,j)$ is obtained using the SLM method as follows:~$S_{app}^H(i,j) = SLM(\mathcal{T}_i(last) , O_j)$.  By integrating $S_{iou}^H(i,j)$ and $S_{app}^H(i,j)$ together, the similarity between  $\mathcal{T}_i$ and the $j$th object $O_j$ in $\mathbb{O}^H$ is calculated as follows:
\vspace{-0.5mm}
\begin{equation}
S^H(i,j) = S_{iou}^H(i,j) + S_{app}^H(i,j). 
\label{Eq_MH}
\end{equation}
\vspace{-0.5mm}
%

\begin{figure}[t]
\centerline{
\includegraphics[width=0.8\linewidth]{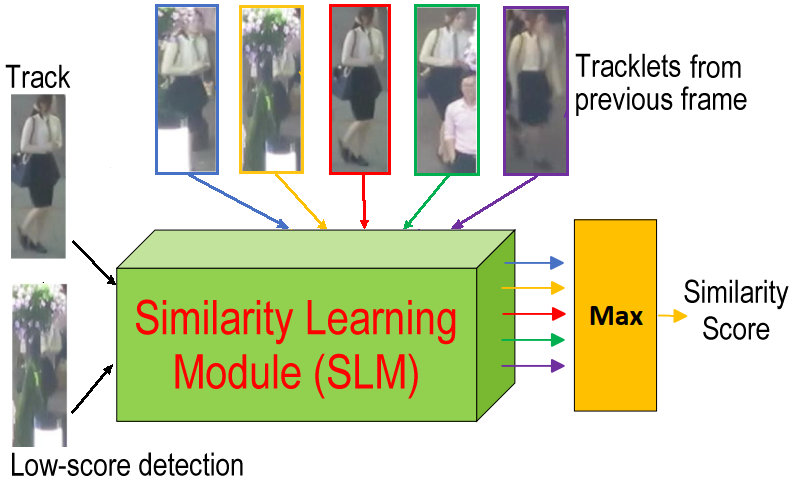}
\vspace{-2mm}
}
\caption{Appearance similarity between low-score detection at the current frame and tracks at the previous frame. Five tracklets compute a similarity score with the low-score detection using SLM. The most similar tracklet is selected, as indicated by the orange arrow in the figure.}
\label{fig:MultiTemplate}
\vspace{-4mm}
\end{figure}

Fig.~\ref{fig:MultiTemplate} shows an example to calculate appearance similarity by the multi-templated SLM. Let $S_{iou}^L(i,j)$ be the IOU similarity matrix between $\mathcal{T}_i(\text{last})$ and the $j$-th object $O_j$ in $\mathbb{O}^L$. Similar to Eq. \eqref{Eq_MH}, the integrated similarity between $\mathcal{T}_i$ and the $j$-th object $O_j$ in $\mathbb{O}^L$ is calculated as:
\begin{equation}
S^L(i,j) = S_{iou}^L(i,j) + S_{app}^L(i,j). 
\end{equation} 

Using $S^H(i,j)$ and $S^L(i,j)$, we initially associate the objects in $\mathbb{O}^H$ with tracklets in $\mathcal{T}_i$. However, due to occlusions or blur, some tracklets in $\mathbb{O}^H$ remain unmatched. To address this issue, we subsequently associate the objects in $\mathbb{O}^L$ with these unmatched tracklets, leading to State-of-The-Art (SoTA) MOT performance. The details of the SMC module are described below:

\noindent{\bf Stage I}:
During the first stage of association, our focus is on finding matches between $\mathbb{O}^H$ and $\mathcal{T}$. We employ the Hungarian algorithm to perform linear assignment using the similarity matrix $S^H(i,j)$. The unmatched objects of $\mathbb{O}^H$ and the unmatched tracks of $\mathcal{T}$ are then placed in $\mathbb{O}_{Remain}^H$ and $\mathcal{T}_{Remain}^H$, respectively.

\begin{table*}[t]
\caption{Comparison against the SoTA methods on the MOT20~\cite{dendorfer2020mot20}  test set. 
}
\label{tab:SoTA:MOT20}
\small
\centering{
\begin{tabular}{c|ccccccc}
\toprule
Method &
  MOTA $\uparrow$ &
  IDF1 $\uparrow$ &
  HOTA $\uparrow$ &
  FN $\downarrow$ &
  FP $\downarrow$ &
  IDs $\downarrow$ & 
  FPS $\uparrow$ \\ \toprule
FairMOT~\cite{zhang2021fairmot} &61.8 &67.3 &54.6 &103,440 &88,901 &5,243 &13.2\\
CSTrack~\cite{CSTrack:TIP:2021} &66.6 &68.6 &54.0 &144,358 &25,404 &3,196 & 4.5\\
TransTrack~\cite{transtrack:arxiv:2020} &65.0 & 59.4 &48.5 & 150,197  & 27,197 & 3,608 &7.2 \\
TransCenter~\cite{TransCenter:TPAMI:2021} &61.9 &50.4 &- &146,347 &45,895 &4,653 &1.0\\
CorrTracker~\cite{CorrTracker:CVPR:2021} &65.2 &69.1 &- &95,855 &79,429 &5,183 &8.5\\
GSDT~\cite{GSDT:ICRA:2021}& 67.1 & 67.5 & 53.6 & 135,409 & 31,913 & 3,131 &0.9\\
SiamMOT~\cite{SiamMOT:AAAI:2021}&67.1 &69.1 &- &-  &- &-  &4.3\\ 
RelationTrack~\cite{RelationTrack:ICIP:2021}&67.2 &70.5 &56.5 &104,597 & 61,134 &4,243 &2.7 \\
SOTMOT~\cite{Zheng_2021_CVPR} & 68.6  & 71.4 &- & 101,154 & 57,064 & 4,209 &8.5\\
MAATrack~\cite{Stadler_2022_WACV}& 73.9 &71.2 & 57.3 &108,744  &24,942 &1,331 & 14.7\\
StrongSORT++~\cite{du2023strongsort} & 73.8  & 77.0  & 62.6 & 117,920  & \textbf{16,632} & \textbf{770} & -\\
OCSORT~\cite{cao2022observation} & 75.7 & 76.3 & 62.4 & 105,894  & 19,067 & 942 & -\\
TransMOT~\cite{TransMot:ICCV:2021} &77.5 &75.2 &-  &\textbf{80,788}  &34,201 &1615 & -\\
ByteTrack~\cite{zhang2021bytetrack}  & 77.8 & 75.2 & 61.3 & 87,594  & 26,249 & 1,223 &17.5\\
BoT-SORT~\cite{aharon2021botsort}  & 77.7  & 76.3 & 62.6 & 86,037  & 22,521 & 1,212 &6.6\\
\textbf{SMILEtrack w/o Re-ID (Ours)} & 78.0 & 76.3 & 63.0 & 86,112  & 23,246 & 1,208 &\textbf{22.9} \\ \midrule
BoT-SORT-R~\cite{aharon2021botsort}  & 77.8  & 77.5 & 63.3 & 88,863  & 24,638 & 1,257 &2.4\\ 
\textbf{SMILEtrack(Ours)} & \textbf{78.2} & \textbf{77.5} & \textbf{63.4} & 85,548  & 24,554 & 1,318 &7.2 \\ 
\bottomrule
\end{tabular}
}
\end{table*}

\noindent{\bf Stage II}:
In the second matching stage, we match the objects in $\mathbb{O}^L$  to the tracklets in $\mathcal{T}_{Remain}^H$. We complete the linear assignment by the Hungarian algorithm with the similarity matrix $S^{L}$. The unmatched objects in $\mathbb{O}^L$ and the unmatched tracks in $\mathcal{T}_{Remain}^H$ are placed in $\mathbb{O}_{Remain}^L$ and $\mathcal{T}_{Remain}^L$.

\begin{figure}[t]
\centerline{
{\footnotesize (a)}
\includegraphics[width=0.95\linewidth]{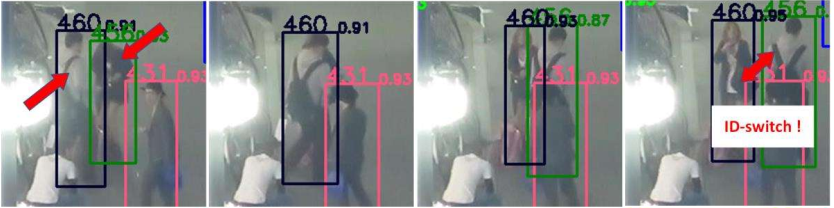}
}
\centerline{
{\footnotesize (b)}
\includegraphics[width=0.95\linewidth]{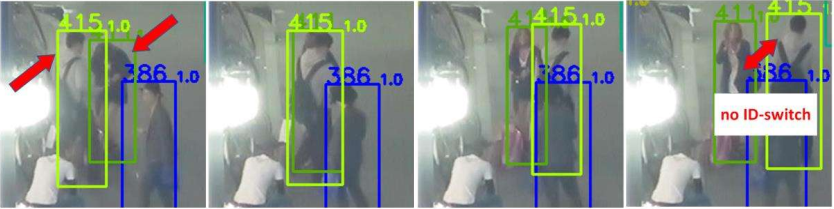}
}
\caption{ 
The use of a GATE function can better handle the occlusion and ID-switch problems in MOT. (a) Results of MOT without using the GATE function.  When the two targets are getting closer and the IOU score is higher than the appearance score, an ID-switch problem happens. (b) Results of MOT using the GATE function. 
}
\label{fig:GATE}
\vspace{-6mm}
\end{figure}

\subsection{The SMC GATE Function} 

To calculate the similarity score, most MOT methods use a weighted sum to combine the IOU and the appearance information to improve the accuracy of data association. However, this method can cause problems when the IOU score is significantly higher than the appearance similarity score between two distinct pedestrians, as they may only overlap, but are not the same. This paper introduces a GATE function in the SMC module to reject a target if its appearance similarity score is low, even when it comes with a high IOU score.

Due to occlusions or lighting changes, objects in $\mathbb{O}_{Remain}^H$ with higher scores may not be matched in the current frames, but their correspondences may potentially be found in future frames. If a target in $\mathbb{O}_{Remain}^H$ passes the GATE function check, the SMC module will generate a new tracklet and add it to $\mathcal{T}$ for further matching. The GATE function uses a threshold $\tau$ to select objects from $\mathbb{O}_{Remain}^H$ if their detection scores are higher than $\tau$ and include them in $\mathcal{T}$ as new tracks for further association. Objects in $\mathbb{O}_{Remain}^H$ with detection scores lower than $\tau$, as well as those in $\mathbb{O}_{Remain}^L$, are considered background and filtered out. It is important to note that tracks in $\mathcal{T}_{Remain}^L$ are deleted if they remain unmatched for more than 30 frames. This GATE function is a novel addition not present in ByteTrack~\cite{zhang2021bytetrack}, and it aims to re-select potential tracks from $\mathbb{O}_{Remain}^H$ to handle challenging scenarios involving severe occlusions. Without this GATE function, ByteTrack cannot determine whether the objects to be matched are seriously occluded or not. Fig.~\ref{fig:GATE} and Table~\ref{tab:abl:bytetrack} shows the advantage of the GATE function.

\section{Experimental Results}
\label{sec:exp}

{\bf Implementation Details.}
Our experiments were conducted on MOT17 and MOT20 benchmarks~\cite{milan2016mot16}, with additional training on datasets~\cite{schoeps2017cvpr,dollarCVPR09peds,milan2016mot16,inproceedings,cloud,EE,xiao2017joint,zheng2017person}. For re-ID models, datasets providing both bounding box location and identity information, such as CalTech~\cite{dollarCVPR09peds}, PRW~\cite{zheng2017person}, and CUHK-SYSU~\cite{xiao2017joint}, were used. Evaluation metrics~\cite{milan2016mot16} included MOTA~\cite{Bernardin2018Metrics}, IDF1~\cite{Ristani2016ECCV}, and HOTA~\cite{HOTA2021}, highlighting detection performance and identity matching. Our detector was initialized on the COCO dataset~\cite{coco} and fine-tuned on MOT datasets, employing data augmentation and an SGD optimizer with cosine annealing. The SMC module introduced a GATE function to manage new tracklets, with key parameters assessed in an ablation study. Additional details regarding the effects of track buffer, template lengths, and patch layout can be found in the supplementary.

\subsection{Evaluation Results}


\begin{table*}[t]
\caption{Ablation analysis of SLM, SMC, and GATE Function (GF) on the MOT17 validation set, compared to the leading ByteTrack~\cite{zhang2021bytetrack} that utilizes the YOLOX~\cite{yolox} detector. The FPS encompasses detection, NMS, re-identification, and data association, excluding image acquisition and video encoding/decoding processes.
\vspace{-2mm}
}
\label{tab:abl:bytetrack}
\centering
\begin{tabular}{l|l|lll|lll|l}
\toprule 
Method  & Detector   & SLM & SMC & GF & MOTA $\uparrow$ & IDF1 $\uparrow$ & IDs $\downarrow$ & FPS $\uparrow$  \\
\toprule 
ByteTrack & YOLOX &     &     &               & 74.1 & 77.0 & 803 & \textbf{9.7}  \\
SMILEtrack & YOLOX  & \checkmark   &     &               & 76.2 & 78.4 & 647 & 8.1  \\
SMILEtrack& YOLOX   & \checkmark   & \checkmark   &               & 76.9 & 79.1 & 594 & 8.0  \\
SMILEtrack& YOLOX   & \checkmark   & \checkmark   & \checkmark             & \textbf{77.5} & \textbf{79.9} & \textbf{554} & 7.5  \\
\hline
SMILEtrack & PRB-Net  &     &     &               & 75.3 & 77.5 & 856 & \textbf{10.2} \\
SMILEtrack & PRB-Net  & \checkmark   &     &               & 77.6 & 79.3 & 601 & 8.5  \\
SMILEtrack & PRB-Net & \checkmark   & \checkmark   &               & 78.2 & 80.2 & 543 & 8.2  \\
SMILEtrack & PRB-Net & \checkmark   & \checkmark   & \checkmark             & \textbf{78.6} & \textbf{80.8} & \textbf{509} & 7.8  \\
\bottomrule
\end{tabular}
\vspace{-4mm}
\end{table*}

\begin{table}[t]
\caption{Effects of patch layouts on performance improvement evaluated on the MOT17 val set.
}
\label{tab:PatchLayout:MOT17}
\setlength\tabcolsep{4.0pt}
\centerline{
\begin{tabular}{c|cccccccc}
\toprule
Method &
  A  &
  B  &
  C  &
  D  &
  E  &
  F  &
  G  &
  H  \\ \toprule
MOTA$\uparrow$& 76.0  & 76.1 & 76.2 & 76.4 & \textbf{76.4} & 76.3  & 76.4  &76.4\\
IDF1$\uparrow$ & 77.4 & 77.6 & 77.7 & 77.9  & \textbf{78.4} &78.2 & 78.4 & 78.3\\
IDs$\downarrow$ &732 &705 &681 &654 &\textbf{624} &645 &633 &630 \\
\bottomrule
\end{tabular}
}
\vspace{-2mm}
\end{table}

\begin{table}[t]
\caption{Performance comparisons of similarity scores on the MOT17 validation set. 
\vspace{-2mm}
}
\label{tab:MOT17:valid}
\hfill
\small
\centering
\centerline{
\begin{tabular}{c|c|ccc}
\toprule 
SMC Stage I & SMC Stage II & MOTA $\uparrow$ & IDF1 $\uparrow$ & IDs $\downarrow$  \\ 
\toprule 
IOU     & IOU     & 76.2 & 74.0 & 731 \\
SLM + IOU & IOU     & \textbf{76.5} & 78.8 & 615 \\
IOU     & SLM+IOU & 76.1 & 73.7 & 740 \\
SLM+IOU & SLM+IOU & 76.4 & 78.5 & 624 \\
SLM+IOU & SLM (Multi)+IOU & \textbf{76.5} & \textbf{78.9} & \textbf{585} \\
\bottomrule
\end{tabular}
}
\vspace{-4mm}
\end{table}

Table~\ref{tab:SoTA:MOT17} presents the evaluation comparisons of our SMILEtrack with State-of-The-Art (SoTA) tracking models on the MOT17 test set, following the evaluation of the MOTChallenge~\cite{milan2016mot16}. All evaluation results were obtained using the official MOTChallenge evaluation website. SMILEtrack outperforms all SoTA methods in several metrics, namely MOTA, IDF1, HOTA, FN, and MT, respectively. Note that the MOT community pays particular attention to the compound metrics MOTA and IDF1. Additionally, in the MOT17 dataset, SMILEtrack is the only method to achieve an IDF1 score higher than 80. ByteTrack~\cite{zhang2021bytetrack} shows high efficiency among SoTA methods, but also exhibits higher false positive and false negative rates. On the other hand, StrongSORT++~\cite{du2023strongsort} achieves the lowest false negatives but with significantly higher false positives. 

Our SMILEtrack is the only one method that achieves a score higher than 80 in the IDF1 metric on the MOT17. ByteTrack is the most efficient among all the SoTA methods but with higher IDs and FN. StrongSORT++ obtains the lowest IDs but with a much higher FN. Table~\ref{tab:SoTA:MOT20} presents comparisons of our SMILEtrack with the SoTA methods on the MOT20 test set. SMILEtrack surpasses all SoTA methods in the MOTA, IDF1, HOTA, and FN metrics on MOT20. ByteTrack remains the most efficient MOT method, while StrongSORT++ achieves the lowest false positives, but still with a much higher FN.




\subsection{Ablation Studies}



\subsubsection{Effects of Patch Layouts}
Different patch arrangements will affect the performance of SLM.  Therefore, the first ablation study aims to investigate the effects of different patch layouts on SLM performance improvements. Fig.~\ref{PatchLayout} shows different types of patch layouts. Table~\ref{tab:PatchLayout:MOT17} shows the effects of different patch layouts on performance improvements evaluated on the MOT17 val set~\cite{milan2016mot16}.  As shown in Fig.~\ref{PatchLayout}, the type-E patch layout outperforms others with respect to the metrics MOTA, IDF1, and IDs.  This paper adopts the type-E patch layout for all performance evaluation and comparison. 

\subsubsection{Re-Identification Strategies.}
Traditional methods primarily rely on IOU to calculate similarity scores~\cite{zhang2021bytetrack}. These methods often fail to track rapidly moving objects effectively due to the lack of appearance matching, leading to an increase in identity switches. As shown in Table~\ref{tab:abl:bytetrack}, SMILEtrack with PRB-Net~\cite{prb} outperforms ByteTrack with YOLOX~\cite{yolox} in terms of all metrics and efficiency because ByteTrack encounters issues regarding re-identification. Noteably, when ByteTrack is incorporated with our methods, such as the SLM, SMC, and GATE functions, its accuracy improves substantially to the level comparable to SMILEtrack, which justifies the effectiveness of SLM, SMC, and GATE.

\subsubsection{Combination of SLM and IOU for different Stages.}
Table~\ref{tab:MOT17:valid} presents performance comparisons for different combinations of IOU and appearance features in Stages I and II of the SMC module. Objects in $\mathbb{O}^H$ generally have higher detection scores and fewer occlusions. Using both the IOU and the appearance features in stage I, SMILEtrack achieves improved MOTA, IDF1, and IDs, as in Table~\ref{tab:MOT17:valid} rows 2 and 3. Objects in $\mathbb{O}^L$ are more prone to occlusions or motion blurring. Thus, SMC Stage II relying solely on IOU yields the best results, as shown in Table~\ref{tab:MOT17:valid} rows 3 and 5. When multiple templates are created in the SLM to address the issue of low detection scores, the best results are obtained by combining SLM (multiple templates) with IOU, as shown in the last row in Table~\ref{tab:MOT17:valid}.



\vspace{-1mm}
\section{Conclusion} 
\vspace{-1mm}

In this paper, we propose SMILEtrack, a Siamese network-like architecture that effectively learns object appearance features for single-camera multiple-object tracking. We introduce the Similarity Matching Cascade (SMC) for bounding box association in each frame, and our experiments demonstrate that SMILEtrack achieves high-performance scores in terms of MOTA, IDF1, IDs, and FPS on the MOT17 and MOT20 datasets.

{\bf Future work.} As SMILEtrack is a Separate Detection and Embedding (SDE) method, it has a slower runtime compared to Joint Detection and Embedding (JDE) methods. In the future, we plan to explore approaches that can improve the efficiency {\em versus} accuracy trade-off in MOT tasks. 

\bibliography{aaai24}

\end{document}